\documentclass[11pt,a4paper]{article}
\usepackage[hyperref]{eacl2021}
\usepackage{times}
\usepackage{latexsym}

\usepackage{microtype}

\usepackage{multirow}
\usepackage{subcaption}
\usepackage{graphicx}
\usepackage{wasysym}

\aclfinalcopy 

\newcommand\LN{\linebreak\noindent}

\usepackage{array}
\newcolumntype{L}{>{\arraybackslash}m{15cm}}

\title{What Went Wrong? \\ Explaining Overall Dialogue Quality through Utterance-Level Impacts}

\author{James D. Finch\footnotemark[1] \\
  Dept. of Computer Science \\
  Emory University \\
  Atlanta, GA, USA \\
  \texttt{jdfinch@emory.edu} \\\And
  Sarah E. Finch\footnotemark[1] \\
  Dept. of Computer Science \\
  Emory University \\
  Atlanta, GA, USA \\
  \texttt{sfillwo@emory.edu} \\\And
  Jinho D. Choi \\
  Dept. of Computer Science \\
  Emory University \\
  Atlanta, GA, USA \\
  \texttt{jinho.choi@emory.edu} \\}

\date{}

\begin{document}
\maketitle

\renewcommand{\thefootnote}{\fnsymbol{footnote}}
\footnotetext[1]{Contributed equally to this work as first authors.}
\renewcommand{\thefootnote}{\arabic{footnote}}

\begin{abstract}

Improving user experience of a dialogue system often requires intensive developer effort to read conversation logs, run statistical analyses, and intuit the relative importance of system shortcomings. 
This paper presents a novel approach to automated analysis of conversation logs that learns the relationship between user-system interactions and overall dialogue quality. 
Unlike prior work on utterance-level quality prediction, our approach learns the impact of each interaction from the overall user rating without utterance-level annotation, allowing resultant model conclusions to be derived on the basis of empirical evidence and at low cost.
Our model identifies interactions that have a strong correlation with the overall dialogue quality in a chatbot setting.\LN
Experiments show that the automated analysis from our model agrees with expert judgments, making this work the first to show that such weakly-supervised learning of utterance-level quality prediction is highly achievable.

\end{abstract}

\section{Introduction}

A typical life cycle of a dialogue system involves many iterative updates where developers improve the system's language understanding capabilities and attempt to increase the overall user engagement.\LN
One of the most challenging aspects of executing these updates is to identify characteristics of the dialogue system that are impacting user experience the most. 
Doing so often involves manually crawling potentially thousands of system logs and designing statistical analyses, both of which are time consuming and unlikely to provide a holistic view of a system's shortcomings.


Inspired by this problem, the presented work investigates the extent to which it is possible to automatically distinguish turns within chat-oriented dialogues that have a negative effect on overall dialogue quality.
The interpretation of dialogue quality is especially difficult in the chat-oriented dialogue setting due to its subjective and multi-faceted objectives. 
System misunderstandings and low user engagement are factors of low quality that are relatively easy to identify, but more subtle factors such as boring responses, awkward topic switches, and individual preferences can also have a substantial effect. 
Furthermore, the practical value of any approach to estimate the quality of individual dialogue turns is highly sensitive to the cost of collecting relevant data. Chatbots, and the settings they are placed in, can differ drastically in both their topics of conversation and interaction styles. And while conversation-level quality labels can be obtained relatively quickly by asking users to provide a rating at the end of a conversation, collecting data with turn-level labels that adequately characterizes a new chatbot or chat setting is an expensive process. 

In this paper we present our dialogue analysis approach, which addresses these challenges by producing quality scores for each utterance in a given conversation dataset using only conversation-level quality ratings. 
Unlike other work that focuses on utterance-level quality prediction using labeled data, our approach involves training a neural model to learn explicit relationships between utterance-level features and conversation quality without the need for costly utterance-level annotations.
We evaluate this approach on two conversation datasets and show high agreement between our model and experts for identifying problematic interactions.
By developing an empirical technique that models the relationship between specific interactions and overall conversation quality, our work has the potential to remove much of the human effort and guesswork involved in dialogue system development.

\section{Related Work}
\label{sec:relatedwork}

Related work has explored techniques for modelling dialogue quality on both the conversation and utterance level.
\newcite{sandbank18} present an approach for classifying low-quality conversations in commercial conversational assistants. 
\newcite{liang20} argued against the feasability of conversation-level quality prediction on a Likert-scale and present a pairwise comparison model instead using methods that compensated for the high noise in user scores. 
\newcite{choi19} presents methods for both predicting user satisfaction and detecting conversation breakdowns at the turn level. 
\newcite{ghazarian20}'s work is similar, predicting utterance-level user engagement.

\newcite{ghazarian20} and \newcite{choi19}'s work is similar to ours, as they build models targeted towards utterance-level quality outcomes.
However, unlike our approach, these works are reliant on costly turn-level annotations: given conversations annotated for quality on the utterance level, their approach is to train a model that can predict utterance quality on unseen conversations within a similar conversation setting. This strategy incurs a substantial cost whenever the training data needs to be updated to fit a novel conversational setting or chatbot. 

To avoid the cost of collecting turn-level labels, our approach is more in line with techniques such as multiple regression analysis, where fitting a model to a dataset is used to explain the relationship between features and some outcome, rather than to predict an outcome for unseen examples. 
In our case, our model can be fit to any dataset of conversations with conversation-level quality labels in order to estimate the quality impact of each utterance on the overall conversation quality.
This approach has a couple advantages over existing work. 
First, collecting utterance-level annotations in a supervised setting is not necessary for our approach as it was for \newcite{choi19} and \newcite{ghazarian20}. 
Second, our model learns empirically-derived relationships between the utterance level and conversation quality, rather than learning to mimic human judgements of utterance level quality irrespective of conversation-level impact.
Given the unreliability of human judgements on conversation quality noted in prior work \cite{liang20}, it is possible that removing human estimations of the relationship between turn-level and conversation-level quality eliminates a source of bias in the model's training objective. 
To our knowledge, no previous work has presented an approach for utterance-level quality estimation that does not require data annotated on the turn level.
\section{Aggregated Regression Analysis}
\label{sec:method}

We utilize a neural network model to accomplish this task of utterance-level quality inference. It learns to assign scores to utterances within a dialogue such that it maximizes its ability to produce correct dialogue-level quality scores from aggregation of these utterance scores. 

Our Aggregated Regression Analysis (ARA) neural model takes as input the entire sequence of utterances for a single dialogue and outputs the predicted dialogue quality $q$ on a continuous scale. Each utterance is first embedded into a continuous space vector representation, producing a sequence of embeddings $(u_1, ..., u_N)$. The rating $r_{i}$ and weight $w_{i}$ of each utterance embedding are then assigned through learned linear transformations of the embedding. The linear transformations are learned independently for the ratings and the weights: 

\begin{equation}
\label{eq:score}
    r_{i} = u_{i}v_{r}^{T} + b_{r}
\end{equation}

\begin{equation}
\label{eq:weight}
    w_{i} = \sigma(u_{i}v_{w}^{T} + b_{w})
\end{equation}

\noindent where parameters $v_{r}$, $ b_{r}$ and $v_{w}$, $b_{w}$ are learned weights and biases for the rating and weight calculations, respectively.

The final dialogue quality prediction $q$ is then calculated as the weighted sum of the utterance ratings as follows, where $N$ is the total number of utterances in a given dialogue:

\begin{equation}
\label{eq:aggreg}
    q = \frac{\sum\limits_{i=0}^N r_{i}w_{i}}{\sum\limits_{i=0}^N w_{i}}
\end{equation}

\begin{figure}[!htbp]
\centering
\includegraphics[width=\columnwidth]{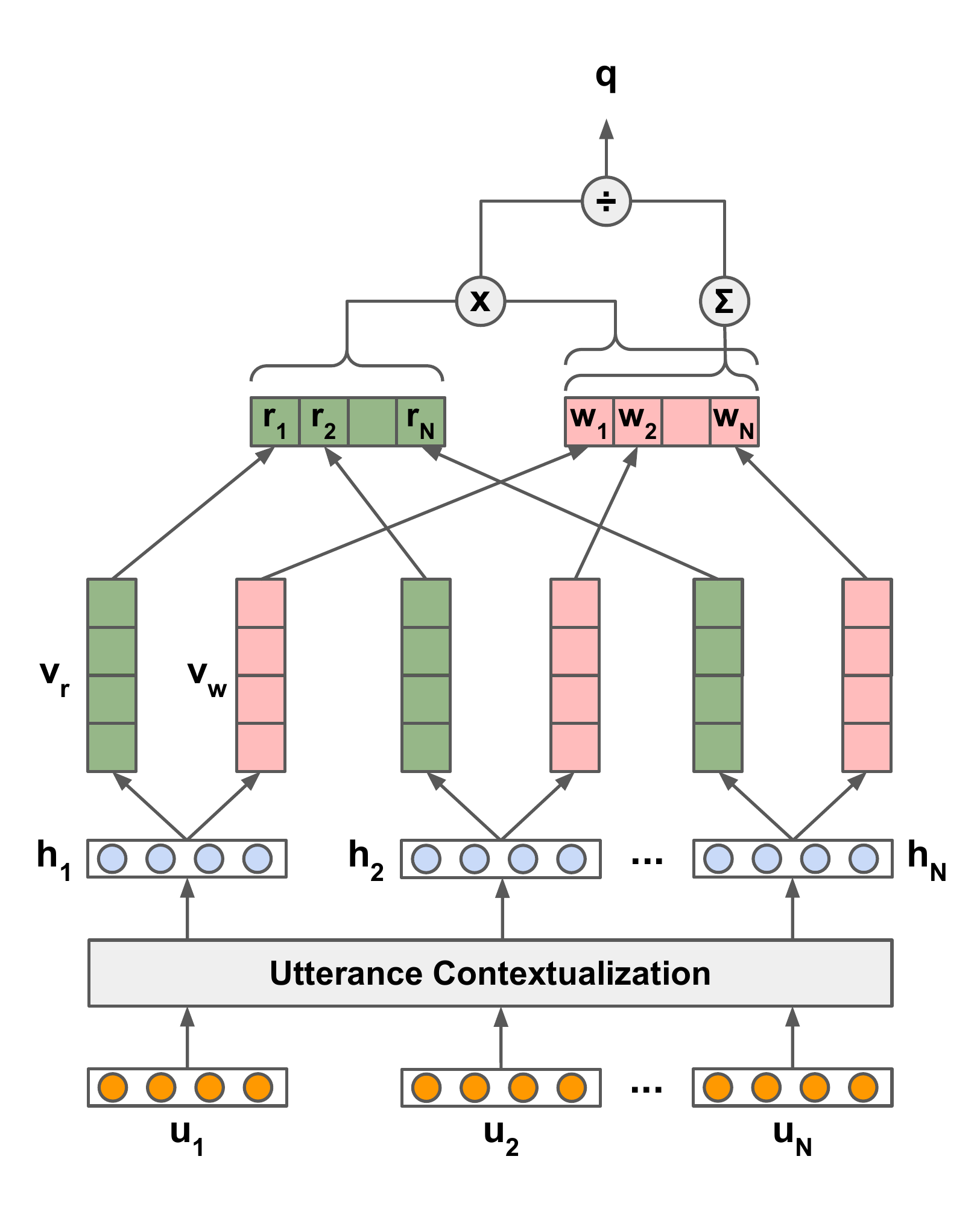}
\caption{Model architecture for predicting dialogue quality by aggregating predicted utterance ratings and weights with contextualization.}
\label{fig:model}
\end{figure}

\noindent We also extend this approach to capture inter-utterance interactions by including an utterance contextualization step. The original utterance embeddings $(u_1, ..., u_N)$ are passed through a contextualization layer to generate a contextualized utterance embedding for each utterance $i$ as $h_{i}$. The resultant rating and weight calculations follow Equations \ref{eq:score} and \ref{eq:weight} with $u_i$ replaced by $h_i$. This final architecture is shown in Figure \ref{fig:model}.

\noindent Although our task is to quantify the relationship between individual utterances and conversation quality, our model is formulated as a conversation-level quality predictor in order to fit to datasets where conversation quality labels are available.
Since the top layers of our architecture enforce this conversation-level prediction to be constructed from individual ratings and weights of each utterance in the conversation, this conversation quality prediction architecture can be applied to the model analysis task that is the focus of this paper.
After fitting this model to a conversation dataset with quality labels, utterance-level effects on quality can be inferred by extracting the ratings and weights of each utterance from the model's intermediate output.
The final inferred impact score $s_i$ of each utterance $u_i$ on the conversation quality is then simply given as a product of the rating and weight:

\begin{equation}
\label{eq:finalscore}
    s_i = r_i * w_i
\end{equation}

\section{Experiment}
\label{sec:experiment}

\subsection{Data}
\label{sec:AP19}

Our task requires dialogue datasets that contain dialogue-level quality ratings. However, to measure the success of inferring utterance-level impacts from these dialogue-level quality ratings, we need to evaluate any approach on this task against dialogues that also contain utterance-level ratings. To our knowledge, there are two publically available datasets that provide both dialogue-level and utterance-level quality ratings: Amazon's Topical-Chat \cite{gopalakrishnan19} and the First Conversational Intelligence Challenge (ConvAI) \cite{burtsev18}.

Although Topical-Chat obtained human annotations on utterance quality, there are several limitations to these annotations that make them unsuitable as an evaluation dataset for our task. For one, the quality of the utterance-level annotations is questionable. Through a preliminary analysis of the dataset, we observed many cases where we could not justify the human annotator’s ratings. In fact, 99\% of the utterance-level annotations received a rating of 4 or 5, and annotators seemed likely to give such a rating with no regard to the coherence of the utterance in its context. Additionally, even if the annotation reliability was not in question, this dramatic bias towards a small subset of the rating classes also leads to the dataset containing a substantial lack of variety in utterance quality, which would make this dataset uninteresting as an evaluation method. For these reasons, we choose to exclude Topical-Chat from our evaluation.

On the other hand, ConvAI contains more distributed utterance-level ratings that did not raise the same quality concerns upon our preliminary analysis. As a result, we include ConvAI in the evaluation of this task. In addition, we augmented a subset of the dialogues collected through our participation in the 2019 Alexa Prize with utterance-level annotations. We use this augmented dataset for evaluation as well. More details on these included datasets are provided next.

\paragraph{ConvAI}

ConvAI was held as a Competition Workshop at NIPS 2017, where 10 teams submitted bots that were able to hold conversations with humans on short text news articles. During evaluation of the bots, humans provided a dialogue quality rating at the conclusion of their conversation, and also had the option of rating each of their conversational partner's turns using an online good/bad indicator. We heldout 200 conversations from the overall dataset as development and test splits. Table \ref{table:convai_stats} shows statistics of the remaining ConvAI conversations ($N$=2,459) used for training and utterance-level evaluation.


\begin{table}[htbp!]
\centering\small 
\begin{tabular}{r|c|c|c|c|c}
\bf Rating & \bf 1 & \bf 2 & \bf 3 & \bf 4 & \bf 5 \\
\hline\hline
\bf Dialogues  & 1100 & 537 & 345 & 308 & 169\\
\bf Proportion  & 44.7 & 21.8 & 14.0 & 12.5 & 6.9\\
\bf Avg. Turns & 5.99 & 12.06 & 12.98 & 12.38 & 13.05 \\
\hline
\bf Good Turns & 406 & 726 & 720 & 1262 & 783 \\
\bf Bad Turns & 1371 & 1168 & 623 & 320 & 111 \\
\end{tabular}
\caption{Statistics for the ConvAI dataset.}
\label{table:convai_stats}
\end{table}

\vspace{-3ex}

\paragraph{AP19}

AP19 consists of dialogues that we collected during the 2019 Amazon Alexa Prize Competition \cite{gabriel20}.
These dialogues occurred between Amazon Alexa users and one socialbot from the 2019 competition. This socialbot was developed to hold in-depth chat-oriented conversations on a wide variety of topics with users, including sports, pets, work, and family.
Dialogues were rated by the user at the conclusion of the dialogue on a voluntary basis by providing a numeric score between one and five in response to the question of how much they would want to talk to this socialbot again.


The AP19 dataset was collected between March and July of 2020. Dialogues with less than 5 utterances were excluded because we observed that such dialogues frequently occurred due to unintentional invocation of the Alexa Prize skill. We heldout 4873 conversations as development and test splits each. Table \ref{table:data_stats} shows statistics of the remaining AP19 conversations ($N$=38,693) used for training and utterance-level evaluation.

\begin{table}[htbp!]
\centering\small 
\begin{tabular}{r|c|c|c|c|c}
\bf Rating & \bf 1 & \bf 2 & \bf 3 & \bf 4 & \bf 5 \\
\hline\hline
\bf Dialogues  & 4785 & 4534 & 5965 & 8385 & 15024\\
\bf Proportion  & 12.4 & 11.7 & 15.4 & 21.7 & 38.8\\
\bf Avg. Turns & 26.04 & 30.22 & 32.80 & 36.57 & 34.98 \\
\end{tabular}
\caption{Statistics for the AP19 dataset.}
\label{table:data_stats}
\end{table}

\subsection{Models}
\label{sec:models}
We trained 3 variants of the ARA model presented in Section \ref{sec:method}, including both the non-contextualized base version and two extensions using different contextualization methods:

\paragraph{Non-Contextualized} This is the base ARA model from Section \ref{sec:method} that does not utilize contextualization of the utterance embeddings.

\paragraph{Order Driven Contextualization} ARA-O extends the base model by using a bidirectional LSTM layer in order to target the importance of utterance order when determining utterance quality. 

\paragraph{Attention Based Contextualization} ARA-A extends the base model by using a self-attention layer to incorporate long-range cross-utterance relationships when determining utterance quality.

\vspace{1ex}
\noindent We also include an additional baseline model:

\paragraph{Non-Aggregated Regression Analysis} NARA is trained on the task of directly predicting utterance scores, instead of a final dialogue score. It employs a heuristic for obtaining target scores for utterances, where each utterance score is equal to the dialogue score in which it occurs. This problem is treated as a sequence regression task, where the model takes as input a sequence of utterances for a given conversation and the utterances are first contextualized using a bidirectional LSTM layer before being passed through a fully connected layer to output the predicted score for each.


\subsection{Model Configurations}
\label{sec:config}

Utterances are embedded using the DistilBERT version of Sentence-BERT (SBERT) by \citet{reimers19}. SBERT is a sentence encoder utilizing a siamese neural network architecture and BERT-based embeddings that has been shown to outperform other methods of encoding sentences on a variety of downstream NLP tasks.

For all ARA-derivative models, hyperparameters were chosen such that they yielded the best-performing models at predicting conversation quality according to Pearson's correlation on a development set of the data. For NARA, the decision was based on performance on utterance-level quality prediction instead. Details on the final configurations are provided in Appendix A.


\subsection{Dialogue Quality Regression Results}

Table \ref{table:dialoguequality} shows the performance of our models on predicting overall dialogue quality. Previous works have shown the difficulty of this task, the noisiness of user ratings, and the low agreement between independent human annotators on the same conversations \cite{liang20, finch20b}. As noted in Section \ref{sec:config}, we used the development performance for hyperparameter tuning only. 

\begin{table}[htbp!]
\centering\small 
\begin{tabular}{r|c|c|c|c}
          & \multicolumn{2}{c|}{\bf AP19} & \multicolumn{2}{c}{\bf ConvAI}\\
\bf Model & \bf Dev & \bf Test & \bf Dev    & \bf Test \\
\hline\hline
\bf ARA  & 0.34 & 0.33 & 0.29 & 0.27 \\
\bf ARA-O & 0.37 & 0.36 & 0.44 & 0.39 \\
\bf ARA-A & 0.36 & 0.35 & 0.34 & 0.25 \\
\hline
\bf NARA  & 0.36 & 0.35 & 0.40 & 0.20 \\
\end{tabular}
\caption{Pearson's $r$ correlations achieved by each model on the dev/test data.}
\label{table:dialoguequality}
\end{table}

\section{Evaluation}
\label{sec:evaluation}

As mentioned in our task formulation (section \ref{sec:task}), evaluating our model is challenging because it requires human judgements about the magnitude and direction of effect of each utterance on the quality of the conversation it is a part of. 
Annotating quality on an utterance level is already a difficult task with much subjectivity involved \cite{higashinaka16}.
Furthermore, asking humans to annotate quality using real numbered values results in arbitrary judgements of magnitude, further complicating a direct evaluation of our task. 
Nevertheless, a model that appropriately assigns utterance scores in a way that explains each utterance's relative impact to the conversation quality should agree substantially with judgements of human experts. 

We address these challenges by presenting two evaluation procedures that approximate the ground truth of our task formulation while framing all human judgements as non-arbitrary decisions with high inter-annotator agreement.
These evaluations focus especially on utterances that have a substantial negative effect on conversation quality, since this case is most in line with the main motivations of our work to automatically identify problematic interactions.

\subsection{Human-Led Evaluation}
\label{sec:HumanLedEval}

Our first evaluation is motivated by the idea that a good model should score interactions with well-known issues lower than most other interactions. To measure this, human annotators provide binary judgements on the quality of each utterance. Given these human annotations, we evaluate our model by treating the binary utterance-level annotations as a partial ordering of the impact of each utterance on the conversation rating against which we can measure the agreement of the model's full ranking of utterances composed of their assigned scores. This agreement is represented by the Concordance Index (C-Index), which measures the extent to which pairs of items in a predicted ranking agree with some ground-truth partial ranking. 

\paragraph{ConvAI} The ConvAI dataset was released with utterance-level ratings by human annotators. Human users provided a binary rating to utterances indicating whether it was good or bad on a voluntary basis. Thus, we use these ratings provided by the human users for the automatic evaluation of our proposed approach on the ConvAI dataset.

\paragraph{AP19} Although AP19 contains dialogue-level quality ratings from human users, it does not contain such ratings for utterances. To this end, we construct an evaluation dataset by asking experts to pick out issues within each dialogue that are likely to cause a lower user rating. We specifically investigate two types of issues: system misunderstandings and user dissatisfaction.
These two issue types are prevalent in our AP19 data and are a frequent cause of conversation quality degradation.

Two of the authors annotated issues in these two categories on 100 randomly sampled dialogues from our AP19 dataset.
Sampling was restricted to only select dialogues with less than a 5 quality rating, since high-rated conversations often have no major problems.
The following guidelines were used to annotate each dialogue:

\begin{enumerate}
    \item Out of the system utterances that signaled a misunderstanding, mark the one that was most likely to negatively impact the user rating.
    \item Out of the user utterances that signaled dissatisfaction, mark the one that most strongly indicated that the user was dissatisfied.
    \item If the severity of two or more system misunderstandings or user dissatisfaction signals cannot be discriminated, mark all of them.
\end{enumerate}


\noindent All conversations in the sample were doubly annotated, with an interannotator agreement (Cohen's kappa) of 0.674 
For evaluation purposes, we ignore utterances that the annotators did not agree on, with the final evaluation containing 158 issue utterances and 3020 non-issue utterances.


\subsubsection{Human-Led Evaluation Results}

\begin{table}[htbp!]
\centering\small 
\begin{tabular}{r|c|c}
\bf Model & \bf AP19 & \bf ConvAI  \\
\hline\hline
\bf NARA & 0.760 & 0.739 \\
\bf ARA  & 0.712 & 0.574 \\
\bf ARA-O & 0.807 & 0.728 \\
\bf ARA-A & 0.638 & 0.599 \\
\end{tabular}
\caption{Average model performance (C-Index) on ranking utterances with respect to human annotations.}
\label{table:cindex_all}
\end{table}

\noindent Table \ref{table:cindex_all} shows the performance of our proposed approach in predicting the correct quality ranking of utterances. These results are based on the average performance across three independent iterations of each model fitted on the datasets. Across all versions of the evaluation, our proposed approach using the order-driven contextualization (ARA-O) displays strong performance, achieving a C-Index of 0.807 on AP19 and 0.728 on ConvAI. For ConvAI, however, ARA-O is slightly outperformed by the NARA baseline. For AP19, the attention-based contextualization (ARA-A) configuration performs much worse, producing performance that is lower than either of the baselines, although it outperforms the base ARA for ConvAI. We observe that ARA-A is prone to significant overfitting which may account for this poor performance.

We also investigated the transferability of the best-performing model trained on one dataset (AP19) to another (ConvAI) shown in Table \ref{table:transfer}. On its own, the model performs poorly on the out-of-domain dialogues which we expected. When finetuned on ConvAI, its performance is comparable to a model trained only on the ConvAI dataset. 


\begin{table}[htbp!]
\centering\small 
\begin{tabular}{r|c}
             & \bf ConvAI \\
\hline\hline
\bf AP19 ARA-O & 0.506 \\
\bf + Finetune & 0.724 \\
\end{tabular}
\caption{Performance of model trained on AP19 when applied to ConvAI.}
\label{table:transfer}
\end{table}



\noindent We include the C-Index scores of the annotators against each other (Table \ref{table:cindex_annot}) in order to provide a strong comparison for our models. Although the difference between our best model and human performance is not too great, there is still a small gap in performance.

\begin{table}[htbp!]
\centering\small 
\begin{tabular}{r|c|c}
             & \bf Annot. 1 & \bf Annot. 2 \\
\hline\hline
\bf Annot. 1 & - & 0.823 \\
\bf Annot. 2 & 0.853 & - \\
\end{tabular}
\caption{Human annotator's agreement (C-Index) on utterance-level issue scores for AP19.}
\label{table:cindex_annot}
\end{table}


\subsection{Model-Led Evaluation}

A major limitation of our human-led evaluation is the restriction to only two kinds of issues. 
Although humans seem capable of judging when well-known problems like misunderstandings and dissatisfaction impact the dialogue quality, there are a number of non-obvious factors that contribute to the user's experience as well.
Data-driven models have a greater potential than humans to uncover these factors, since they can make inferences based on information aggregated across thousands of conversations rather than on biased intuition.
However, given that such a model outputs some estimation of an utterance's impact on conversation quality, that output should be verifiable by a human and agree substantially with expert judgement.

To account for a broader spectrum of factors that may affect conversation quality, we conducted an additional evaluation that asks experts to judge pairs of interactions, where one interaction was assigned a low score by the model and one was assigned a high score. 
We choose the 5th percentile as a cutoff to distinguish issues (low-scored) from non-issues (high-scored), and randomly sample one utterance from each side of the distribution to construct pairs.
Sampling is not done at uniform random, since we observed in a pilot evaluation that this results in many similar samples that are often paraphrases of one another. 
Instead, we run $k$-means clustering on the pretrained SBERT embeddings of the bottom 5\%-scored utterances and restrict our sampling to the $k$ utterances closest to the centroid of each cluster.
We chose $k$ to be 1\% of the total number of issue utterances being clustered, $k=683$.
This procedure ensures a high degree of variety among chosen samples, increasing the robustness of the evaluation.

Each sampled issue from the clustering procedure is paired with a non-issue drawn uniform-randomly from the upper 95\%-scored utterances. 
Each pair of utterances is then presented to the human judge with random intra-pair order, so the judge is blind to which utterance was scored lower by the model.
We allow judges to see two preceding utterances and one following utterances from the full conversation when viewing each example.
Providing a context window of this size was done in order to focus the judgement on a specific interaction of the dialogue, while still providing sufficient background to interpret the sampled utterance correctly.
Given expert judgements on these pairs, model accuracy is calculated as the proportion of times expert judgements agreed with model score assignments. We retrieve 300 pairs for evaluation from our AP19 dataset and two of the authors performed the evaluation. 

\subsubsection{Model-Led Evaluation Results}

We conducted the Model-Led Evaluation on ARA-O for our AP19 dataset, since it outperformed all other models in the initial Human-Led Evaluation. Table \ref{table:acc_all} presents the model's ability to discriminate quality ratings between utterances. Our model is able to achieve an accuracy of 0.775 on average.

\begin{table}[htbp!]
\centering\small 
\begin{tabular}{r|c}
 & \bf Accuracy \\
\hline\hline
\bf Annot. 1 & 0.77 \\
\bf Annot. 2 & 0.78 \\
\hline
\bf Average  & 0.775 \\
\end{tabular}
\caption{Model accuracy at identifying low-quality utterances. \textbf{Annot \#} is relative to individual annotators' selections. \textbf{Average} takes the average of the annotators' accuracy results.}
\label{table:acc_all}
\end{table}

\section{Error Analysis}
\label{sec:errors}

\begin{table}[htbp!]
\centering\small 
\begin{tabular}{r|c|c}
\bf Mistake Type & \bf Frequency & \bf Percentage \\
\hline\hline
Both Issues & 37 & 53\%\\
Both Non-issues & 24 & 34\% \\
Flipped Assignment & 9 & 13\% \\
\end{tabular}
\caption{Distribution of the three mistake types.}
\label{table:errors}
\end{table}

\noindent We conducted an error analysis to determine the frequently occurring mistakes of the proposed approach. To do this, we manually reviewed the 70 utterance pairs and their contexts that resulted in disagreement between the system and at least one annotator in the System-Led Evaluation. We created 3 categories that captured the underlying problems that caused the disagreement. \textit{Both Issues} contains cases where the non-issue selection was as problematic as its issue selection. \textit{Both Non-issues} represents cases where the issue selection was not interpretable as a problematic conversation exchange. \textit{Flipped Assignment} captures the cases where the non-issue selection was more problematic than the issue selection.  Table \ref{table:errors} illustrates the distribution of identified mistake categories across the 70 pairs, with examples shown in Table \ref{table:mistake-examples}. 


\begin{table}[htbp!]
\centering\small 
\begin{tabular}{r|c|c}
\bf Dialogue Pattern & \bf Issue Prop. & \bf Nonissue Prop. \\
\hline\hline
Stop Request & 20\% & 7\%\\
Irrelevant Topic & 16\% & 6\% \\
Topic Switch Request & 13\% & 3\% \\
Unengaging Handoff & 11\% & 3\% \\
\end{tabular}
\caption{Distribution of the dialogue patterns within system-selected issue contexts and nonissue contexts.}
\label{table:patterns}
\end{table}

\begin{table*}[htbp!]
\centering\small 

\begin{subtable}{\textwidth}
\centering\small
\begin{tabular}{L}
\caption*{\bf Example of system-chosen issue utterance:}
\textbf{System:} Oh, that is unfortunate, you should definitely try it sometime, V R is slowly becoming the future of gaming. Another emerging area is augmented reality, or A R .  what do you think of augmented reality? \\
\textbf{User:} i don't know what that is \\
\textbf{System:} \textit{You don't have any ideas? That is ok. It is still a relatively new thing, so it may be hard to form an opinion. If you were to play a mobile game using A R, what would make it the most fun, do you think?} \\
\textbf{User:} what is a r. \\
\end{tabular}
\label{table:correct}
\vspace{1.0ex}
\end{subtable}


\begin{subtable}{\textwidth}
\centering\small 
\begin{tabular}{L}
\caption*{\bf Example of system-chosen issue utterance with \textit{Topic Switch Request}:}
\textbf{System:} Yeah.  You know I hear the grocery stores in affected areas have lots of empty shelves right now. I'm not sure how necessary it is, but have you been stocking up on supplies?  \\
\textbf{User:} yeah  \\
\textbf{System:} \textit{I guess you can never be too prepared.  You know, sometimes I think, if everyone just stayed home and avoided public areas for a couple weeks, the whole thing could just blow over. But realistically, I think slowing the virus down so that everyone doesn't get sick at once is the best we can hope for.} \\
\textbf{User:} let's talk about the drivers  \\
\end{tabular}
\label{table:switch}
\vspace{1.0ex}
\end{subtable}

\begin{subtable}{\textwidth}
\centering\small 
\begin{tabular}{L}
\caption*{\bf Example of system-chosen non-issue utterance to which annotators disagreed:}
\textbf{System:} Yeah, that is true. There are probably many difficult pieces in the puzzle of teleportation, but we can leave that up to the scientists.      I cannot seem to remember any movies that involve teleportation, can you? \\
\textbf{User:} i don't know \\
\textbf{System:} \textit{Good one. I will have to remember that. Ok. tell me, what was the last movie you liked?}  \\
 \textbf{User:} i like joker \\
\end{tabular}
\label{table:fneg}
\vspace{0.5ex}
\end{subtable}

\caption{Examples illustrating highlights and shortcomings of the proposed approach, where the utterances assigned a low quality are indicated by italics. These examples are similar to interactions seen in the Alexa Prize logs (conversations have been modified to protect customer privacy).}
\label{table:mistake-examples}
\end{table*}

\noindent Upon closer analysis of these disagreed pairs, we observed some dialogue patterns that had high correlation with the system assigning a low score to an utterance as shown in Table \ref{table:patterns}. Using a context window of size 2 around the selected issue, we observed the following patterns:
\paragraph{Stop Request} the selected issue was within one turn of the user requesting the conversation to be over.

\paragraph{Irrelevant Topic} the selected issue occurred near the user sharing that they did not have a particular trait (e.g. not in school, no job, no pets, etc.).

\paragraph{Topic Switch Request} the selected issue was within one turn of the user requesting a new topic.

\paragraph{Unengaging Handoff} the selected issue was within one turn of the system transitioning out of one topic using one of three unengaging statements, such as \textit{Ok well I find it inspiring that we have so many different types of music to listen to}.

\vspace{1.5ex}

\noindent The frequency of these dialogue patterns is shown in Table \ref{table:patterns} and examples illustrating these dialogue patterns can be found in Table \ref{table:mistake-examples}. 

These dialogue patterns were present in many of the pairs in the \textit{Both Non-issues} class. In isolation, the instances of these dialogue patterns do not seem indicative of a problematic conversational interaction; however, when taken as an emergent pattern, they begin to illuminate types of interactions that may be subtly indicative of poor user experience. For instance, there would be no reason to request a new topic or to end the conversation if the user was enjoying themselves, thus these actions are a good indication of the user being dissatisfied. Similarly, our socialbot was better equipped to talk about certain life traits, such as being in school or having a pet, and the analogous conversation for those users who did not fit those categories were not as strong. It could be the case that the model's selection of such utterances was not incorrect and rather it was reaching appropriate conclusions based on a large volume of information it was working with, whereas the human annotators were unaware of this, although it is impossible to verify this through our evaluation procedure. 

On the other hand, for the \textit{Stop Request} category, it could also be the case that the model is over-relying on the stop utterances as a negative signal. Knowing the end of the conversation provides the model with the opportunity to make any last-second corrections to achieve the dialogue quality it needs to assign for the current conversation.

\section{Discussion}
\label{sec:discussion}

By achieving a C-Index score of 0.80 and 0.72 on the AP19 and ConvAI datasets respectively, our approach demonstrates the feasability of inferring utterance-level quality information from aggregation over a dataset of rated conversations. Since C-Index is a generalization of the AUC curve, random decision-making would result in a performance of 0.5, thus we obtain improvements of over 20\% against random.

The success of our proposed approach on two datasets also demonstrates its applicability across different dialogue settings. A key difference between the AP19 and ConvAI datasets is their rating distributions: ConvAI was more right-skewed whereas AP19 was more left-skewed. In addition, the ConvAI dataset employs a text-based interface, includes both human-machine and human-human dialogues, and the dialogues were less socially-oriented. This suggests that our approach to utterance-level explainability is not constrained to only one dialogue system configuration.

It is worth noting that the NARA baseline slightly outperforms the presented approach on the ConvAI dataset. This suggests that parsing through all of the variability present in noisy quality ratings across users to determine utterance-level correlations is potentially challenging when given a small dataset. It may be more suitable to use the conversation rating as a heuristic utterance-level score target in these cases. However, the difference between the proposed approach and this baseline is not too great, so it is difficult to make a strong conclusion.


\section{Conclusion}

Our approach and evaluation results demonstrate the feasibility of an automatic approach for dialogue quality analysis. 
By training a model to learn the relationship between utterance-level features and conversation-level outcomes, it is possible to obtain an empirically-derived ranking of interactions based on whether they relate to positive or negative conversation quality. 
This work has obvious applications in dialogue system development, mitigating the high developer effort involved in manual analysis and human bias in determining the relative importance of system issues.

\section*{Acknowledgments}
We gratefully acknowledge the support of the Alexa Prize Socialbot Grand Challenge 3.
Any contents in this material are those of the authors and do not necessarily reflect the views of the Alexa Prize.

\bibliography{eacl2021}
\bibliographystyle{acl_natbib}

\end{document}